\documentclass[sigconf]{acmart}

\AtBeginDocument{%
  \providecommand\BibTeX{{%
    \normalfont B\kern-0.5em{\scshape i\kern-0.25em b}\kern-0.8em\TeX}}}
\usepackage[ruled,linesnumbered]{algorithm2e}
\usepackage{listings}
\usepackage{multirow}
\usepackage{arydshln}
\usepackage{amsmath}
\usepackage{caption}
\usepackage{subcaption}
\usepackage{mathtools}
\usepackage{bbm}
\usepackage{graphicx} 
\usepackage{float}
\usepackage{booktabs}
\usepackage{pgfplots}
\usepackage{tikz}
\usepgfplotslibrary{groupplots}
\newcommand{\argmax}{\mathop{\mathrm{argmax}}}          

\floatstyle{plaintop}
\restylefloat{table}

\setcopyright{acmcopyright}
\copyrightyear{2018}
\acmYear{2018}
\acmDOI{10.1145/1122445.1122456}
\acmConference[Woodstock '18]{Woodstock '18: ACM Symposium on Neural
  Gaze Detection}{June 03--05, 2018}{Woodstock, NY}
\acmBooktitle{Woodstock '18: ACM Symposium on Neural Gaze Detection,
  June 03--05, 2018, Woodstock, NY}
\acmPrice{15.00}
\acmISBN{978-1-4503-XXXX-X/18/06}

\begin{document}

\title{Choosing the Best of Both Worlds: \\
Diverse and Novel Recommendations through Multi-Objective Reinforcement Learning}

\author{Du\u{s}an Stamenkovi\'{c}}
\affiliation{%
  \institution{University of Novi Sad, Serbia}
  \country{}}
\authornote{Full affiliation is Faculty of Sciences, University of Novi Sad. This work is done when taking an internship in Telefonica Research, Spain, and it was supported by \hyperlink{https://www.c4iiot.eu/}{C4IIOT}.}
\email{dusan.stamenkovic@dmi.uns.ac.rs}

\author{Alexandros Karatzoglou}
\affiliation{%
  \institution{Google Research, United Kingdom}
  \country{}}
\email{alexkz@google.com}

\author{Ioannis Arapakis}
\affiliation{%
  \institution{Telefonica Research, Spain}
  \country{}}
\email{ioannis.arapakis@telefonica.com}

\author{Xin Xin}
\affiliation{%
  \institution{Shandong University, China}
  \country{}}
\email{xinxin@sdu.edu.cn}

\author{Kleomenis Katevas}
\affiliation{%
  \institution{Telefonica Research, Spain}
  \country{}}
\email{kleomenis.katevas@telefonica.com}
 
\begin{abstract}
Since the inception of Recommender Systems (RS), the accuracy of the recommendations in terms of relevance has been the golden criterion for evaluating the quality of RS algorithms. However, by focusing on item relevance, one pays a significant price in terms of other important metrics: users get stuck in a "filter bubble" and their array of options is significantly reduced, hence degrading the quality of the user experience and leading to churn. Recommendation, and in particular session-based/sequential recommendation, is a complex task with multiple - and often \textit{conflicting} objectives - that existing state-of-the-art approaches fail to address. 

In this work, we take on the aforementioned challenge and introduce \textit{Scalarized Multi-Objective Reinforcement Learning (SMORL)} for the RS setting, a novel Reinforcement Learning (RL) framework that can effectively address multi-objective recommendation tasks. The proposed SMORL agent augments standard recommendation models with additional RL layers that enforce it to simultaneously satisfy three principal objectives: \textit{accuracy, diversity}, and \textit{novelty} of recommendations. We integrate this framework with four state-of-the-art session-based recommendation models and compare it with a single-objective RL agent that only focuses on accuracy. Our experimental results on two real-world datasets reveal a substantial increase in aggregate diversity, a moderate increase in accuracy, reduced repetitiveness of recommendations, and demonstrate the importance of reinforcing diversity and novelty as complementary objectives.
\end{abstract}



\maketitle

\section{Introduction}

Whether in the context of entertainment, social networking or e-commerce, the sheer number of choices that modern Web users face nowadays can be overwhelming. Contrary to the common belief that more options are always better, selections made from large assortments can lead to a choice overload~\cite{iyengar1999rethinking} and impair users' capacity for rational decision making. Simply put, when presented with large array situations (e.g., limitless products to purchase from or media content to consume), users are at higher risk of feeling like they made the wrong decision and experience regret, which can degrade the quality of experience with an online service or platform. The problem becomes further aggravated when one is inclined to consider the costs and benefits of all alternative options.

Recommender Systems (RS) alleviate this \textit{paradox of choice}~\cite{schwartz2004paradox} by acting as \textit{second-order strategies}~\cite{sunstein1999second} that facilitate access to relevant information and improve the browsing experience~\cite{hu2018reinforcement, yuan2020}. Hence, in settings where the abundance of options can result in unsatisfying choices or, even worst, abandonment, the user experience is ultimately determined by the RS capacity to filter irrelevant content and recommend only items regarded as desirable. So far, the main focus of the research community in the area of RS has been placed on designing algorithms that can identify and recommend relevant content. However, while doing so, they tend to optimise (for the most part) mainstream metrics such as accuracy, at the expense of other content-derived qualitative aspects. In this work, the term ``accuracy'' denotes the performance of the RS in terms of ranking relevant items in the offline test set, and it should not be mistaken for accuracy in classification tasks.

In recent years, diversity and novelty of recommendations have been recognized as important factors for promoting user engagement, since recommending a diverse set of relevant items is more likely to satisfy users' variable needs. For example,~\citet{hu2011helping} report a strong positive correlation between diversity of recommendations and ease of use, perceived usefulness, and intentions to use the system. Therefore, a RS that suggests strictly relevant items to a user who just purchased an espresso machine will, most likely, end up recommending more coffee machines, while the preferred set of recommendations would include coffee mugs, cleaning equipment, coffee beans, so to speak. In the former case, users will get to interact only with a small subspace of the available item space~\cite{pariser2011filter} and, according to the ``law of diminishing marginal returns'', the utility of the recommendations will eventually degrade as users are exposed to similar content, over and over again. 

Session-based recommendation has been introduced as an alternative, industry-relevant approach to RS. In session-based recommendation, a sequential model (e.g., a RNN \cite{hidasi2015session} or a transformer \cite{kang2018self, vaswani2017attention}) is trained in a self-supervised fashion to predict the next item in the sequence itself, instead of some ``external'' labels~\cite{hidasi2015session, kang2018self, yuan2020}. This training process was inspired by language modelling tasks where, given a word sequence input, the language model predicts the most likely word to appear next ~\cite{mikolov2013efficient}. However, this training method can also produce sub-optimal recommendations, since the loss function is defined purely by the mismatch between model predictions and the actual items in the sequence. Models trained under such a loss function focus only on matching the sequence of clicks a user may generate, while forfeiting other desirable objectives. For example, a service provider may want to promote recommendations that will converge to purchases, increase user satisfaction, diversify user-item interactions and promote long-term engagement. Nevertheless, in order to optimize an RS towards said objectives, one needs to capture them with a differentiable function, which is not a trivial task. Therefore, the use of multi-objective optimization (MOO) is heavily limited in areas where important objectives can only be presented in a form of non-differentiable functions/metrics.

Diversity and novelty of recommended item lists are correlated with increased diversity of sales \cite{fleder2007recommender}, and address the ``winner-takes-all'' problem by recommending less popular items from the so-called ``long-tail'. An item from a diverse recommendation list is more likely to be novel, i.e., an item that the user would not normally interact with. This is supported by prior work, which suggests that most users appreciate novel and less popular recommendations~\cite{lathia2010temporal, zhang2012auralist}. Recommendation models trained with simple supervised learning may encounter difficulties in addressing the above recommendation expectations and the multi-objective nature of many online tasks.

To address the current challenges, we expand on the idea of utilising RL in the RS setting and introduce a \textit{Scalarized Multi-Objective RL} (SMORL) approach. SMORL uses a single RL agent to simultaneously satisfy three, potentially conflicting, objectives: i) promote clicks, ii) diversify the set of recommendations, and iii) introduce novel items, while at the same time optimising for relevance. The model focuses on the chosen rewards while maintaining high relevance ranking performance. More specifically, given a generative sequential or session-based recommendation model, the (final) hidden state of the model can be seen as it’s output layer, since it is multiplied with the last (dense softmax) layer to generate the recommendations~\cite{hidasi2015session, kang2018self, yuan2020}. We augment these models with multiple final output layers. The conventional self-supervised head, is trained with the cross-entropy loss to perform ranking, while the SMORL part is simultaneously trained to modify the rankings of the self-supervised head. The RL heads can be seen as regularizers that introduce more diverse and novel recommendations, while the ranking-based supervised head can provide more robust learning signals (including negative signals) for parameter updates. One of the main advantages of using MORL instead of MOO in the context of RS is the possibility of using non-differentiable functions for reward system that the RL agent uses to regularize the base model.

Previous attempts of balancing accuracy with diversity and novelty included re-ranking of the final set of recommendations or training of multiple models and the use of genetic algorithms to aggregate those models~\cite{ribeiro2012pareto}, whereas our approach relies on training a single model and using the SMORL framework to balance the principal recommendation objectives. We argue that this framework can be easily extrapolated to other domains such as music, video, and news recommendations (by using embedding systems~\cite{budhrani2020music2vec, ma2019news2vec}), where diversity and novelty are high-value metrics. In summary, our work makes the following contributions:
\begin{itemize}
  \item We devise a novel diversity reward that utilises the item embedding space.
  \item We devise a novel metric for evaluation of RS that measures repetitiveness of recommendations.
  \item To the best of our knowledge, we apply Multi-Objective Reinforcement Learning (MORL) in the setting of RS for the first time and explore some of the many possibilities and future research directions that this approach offers. 
  \item We introduce SMORL that drives the self-supervised RS to produce more accurate, diverse and novel recommendations. We integrate four state-of-the-art recommendation models into the proposed framework.
  \item We conduct experiments on two real-world e-commerce datasets and demonstrate less repetitive recommendations sets, significant improvements in aggregate diversity metrics (up to 20\%), all while maintaining, or even improving accuracy for all four state-of-the-art models.
\end{itemize}
  
\section{Related Work}

Several deep learning-based approaches that model the user interaction sequences effectively have been proposed for RS. \citet{hidasi2015session} used gated recurrent units (GRU) \cite{cho2014properties} to model user sessions, while \citet{tang2018personalized} and \citet{yuan2020} used convolutional neural networks (CNN) to capture sequential signals. \citet{kang2018self} exploited the well-known Transformer \cite{vaswani2017attention} in the field of sequential recommendation, with promising results. All of these models can serve as the base model whose input is a sequence of user-item interactions and the output is a latent representation that describes the corresponding user state.

Several attempts to use RL for RS have also been made. In the off-policy setting, \citet{chen2019top} and \citet{zhao2018recommendations} proposed the use of propensity scores to perform off-policy correction, but with training difficulties due to high-variance. Model-based RL approaches \cite{chen2019generative, shang2019environment, zou2019reinforcement} first build a model to simulate the environment, in order to avoid any issues with off-policy training. However, these two-stage approaches depend heavily on the accuracy of the simulator. \citet{xin2020self} introduced SQN and SAC, two self-supervised RL frameworks for RS that augment the recommendation model with two output layers (heads). First head is based on the cross-entropy supervised loss, while the other RL head is based on the Double Q-learning \cite{hasselt2010double}. Although SQN and SAC improve performance, they only increase accuracy by promoting clicks and purchases that a user might make. However, an accurate RS is not necessarily a useful one: real value lies in suggesting items that users would likely not discover for themselves, that is, in the novelty and diversity of recommendations \cite{herlocker2004evaluating}. Improving accuracy typically decreases diversity and novelty, which can occur when RL is deployed to regularize session-based RS (see discussion in Section \ref{experiments}). A decrease of aggregate diversity can impact the user experience and satisfaction with the RS \cite{hu2011helping}. \citet{anderson2020algorithmic} also report that current recommendations discourage diverse user-item interactions.
    
Diversifying recommendations and introducing novel recommendations were recently recognized as important factors for improving RS. Early efforts focused on post-processing methods that aimed to balance accuracy and diversity \cite{qin2013promoting, ashkan2015optimal, sha2016framework}. In order to mitigate issues with significant cumulative loss on the ranking function, personalized ranking methods were proposed \cite{cheng2017learning}. \citet{chen2017fast} tried to address the issues of post-processing methods that consider only pairwise measures of diversity and ignore correlations between items, by proposing the probabilistic model Determinantal Point Process \cite{lavancier2015determinantal} that captures the correlation between items using a kernel matrix. Once this matrix is learned, many sampling techniques can generate a diverse set of items \cite{wilhelm2018practical, chen2017fast, warlop2019tensorized}. These models achieve a trade-off between accuracy and diversity at best. On the other hand, SMORL significantly increases diversity and slightly improves the accuracy.
  
In the RL setting, \citet{zheng2018drn} focused on exploration-exploitation strategies for promoting diversity, by randomly choosing random item candidates in the neighborhood of the current recommended item. \citet{hansen2021shifting} proposed a RL sampling-based ranker that produces a ranked list of diverse items. This model is a simple ranker and the model itself doesn't learn to produce diverse set of items, while the learning process utilizes the REINFORCE algorithm \cite{williams1992simple} which is known to suffer from high-variance. Finally, prior attempts to optimize multiple objectives in the setting of RS relied on Pareto-Optimization using grid search \cite{ribeiro2012pareto} or multi-gradient descent \cite{milojkovic2019multi}. However, by definition, one Pareto optimal solutions is not necessarily better than other Pareto optimal solution with respect to all objectives. 

\section{Multi-Objective RL for RS}
  Let $\mathcal{I}$ denote the whole item set, then a user-item interaction sequence can be represented as $\mathbf{x}_{1:t}$ = \{$x_1, x_2, ..., x_{t-1}, x_t$\}, where $x_i \in \mathcal{I}(0 < i \leq t)$ is the index of the interacted\footnote{In a real world scenario there may be different kinds of interactions. For instance, in e-commerce, the interactions can be clicks, purchases, basket additions, and so on. In music recommendation, the interactions can be characterized by the play time of a song, the number of times a song was listened, etc.} item at timestamp $i$. The goal of next item recommendation is recommending the item $\mathbf{x}_{t+1}$ to users that will best suit their current interests, given the sequence of previous interactions $\mathbf{x}_{1:t}$.
  
  From the perspective of MORL, the next item recommendation task can be formulated as a Multi-Objective Markov Decision Process (MOMDP) \cite{white1980solution}, in which the recommendation agent interacts with the environments $\mathcal{E}$ (users) by sequentially recommending items to maximize the discounted cumulative rewards. The MOMDP can be defined by tuples of$(\mathbf{\mathcal{S}},\mathbf{\mathcal{A}},\textbf{P},\text{\textbf{R}},\rho_0,\gamma)$ where:
  \begin{itemize}
      \item $\mathbf{\mathcal{S}}:$ a continuous state space that describes the user state. The state of the user at timestamp $t$ can be defined as $\mathbf{s}_t = G(\mathbf{x}_{1:t}) \in \mathcal{S} (t>0)$, where $G$ is a sequential model that will be discussed in Section~\ref{section:method}.
      \item $\mathbf{\mathcal{A}}:$ a discrete action space that contains candidate items. The action $a$ of the agent is to recommend the selected item. In the offline RL setting, we either extract the action at timestamp $t$ from the user-item interaction, i.e., $a_t = x_{t+1}$, or by setting it to a top prediction obtained from the self-supervised layer. The ``goodness'' of a state-action pair $(\mathbf{s}_{t}, a_t)$ is described by its multi-objective Q-value function $\mathbf{Q}(\mathbf{s}_{t}, a_t)$. 
      \item $\textbf{P}: \mathbf{\mathcal{S}} \times \mathbf{\mathcal{A}} \times \mathbf{\mathcal{S}} \rightarrow \mathbb{R}$ is the state transition probability $p(\mathbf{s}_{t+1}|\mathbf{s}_t, a_t)$, i.e., a probability of state transition from $\mathbf{s}_t$ to $\mathbf{s}_{t+1}$ when agent takes action $a_t$. 
      \item $\text{\textbf{R}}: \mathbf{\mathcal{S}} \times \mathbf{\mathcal{A}} \mapsto \mathbb{R}^m$ is the vector-valued reward function\footnote{Each component corresponds to one objective.}, where $\mathbf{r}(\mathbf{s}, a)$ denotes the immediate reward by taking action $a$ at state $\mathbf{s}$.
      \item $\rho_0$ is the initial state distribution with $\mathbf{s}_0 \sim \rho_0$.
      \item $\gamma \in [0,1]$ is the discount factor for future rewards. For $\gamma = 0$, the agent only considers the immediate reward, while for $\gamma = 1$, all future rewards are regarded fully except the one of the current action.
  \end{itemize}
  The goal of the MORL agent is to find a solution to a MOMDP in a form of target policy $\pi_\theta(a|\mathbf{s})$ so that sampling trajectories according to $\pi_\theta(a|\mathbf{s})$ would lead to the maximum expected cumulative reward:
  \begin{equation*}
      \max_{\pi_\theta}\mathbb{E}_{\tau \sim \pi_\theta}\bigl[ f \bigl( \textbf{R}(\tau)\bigr)\bigr],\; \text{where} \; \textbf{R}(\tau)= \sum_{t=0}^{|\tau|}{\gamma^t \mathbf{r}(\mathbf{s}_t, a_t)}
  \end{equation*}
  where $f:\mathbb{R}^m \mapsto \mathbb{R}$ is a scalarization function, while $\theta \in \mathbb{R}^d$ denotes the policy parameters. The expectation is taken over trajectories $\tau = (\mathbf{s}_0, a_0, \mathbf{s}_1, a_1...)$, obtained by performing actions according to the target policy: $\mathbf{s}_0 \sim \rho_0, a_t \sim \pi_\theta(\cdot|\mathbf{s}_t), \mathbf{s}_{t+1} \sim \mathbf{P}(\cdot|\mathbf{s}_t, a_t)$.
  
  A scalarization function $f$ maps the multi-objective Q-values $\mathbf{Q}(\mathbf{s}_t, a_t)$ and a reward function $\mathbf{r}(\mathbf{s}_t, a_t)$ to a scalar value, i.e., the user utility. In this paper, we focus on linear $f$; each objective $i$ is given an importance, i.e. weight $w_i, i=1,...,m$ such that the scalarization function becomes $f_{\mathbf{w}}(\mathbf{x}) = \mathbf{w}^\top \mathbf{x}$, where $\mathbf{w} = [w_1,..,w_m]$. 

\begin{figure*}[!t]
\centering
\includegraphics[width=0.89\textwidth]{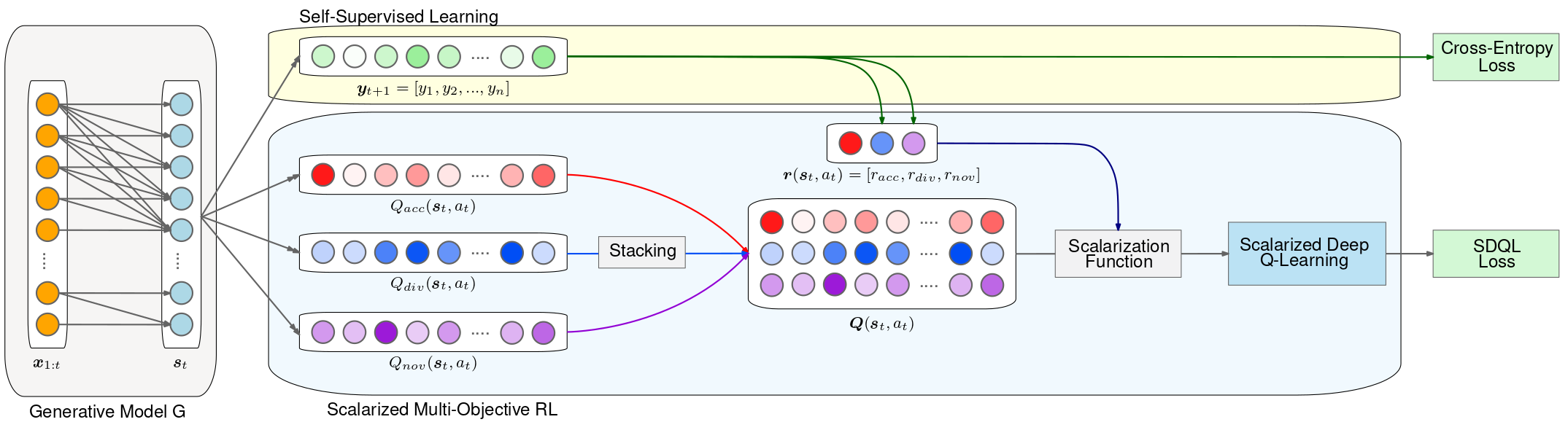}
\vspace{-0.75em}
\caption{The SMORL for Recommender Systems (SMORL4RS) training routine for sequential or session-based RS. Generative model G maps user-item interaction sequence $\mathbf{x}_{1:t}$ to the latent state $\mathbf{s}_t$. With the use of fully-connected layers, $\mathbf{s}_t$ is mapped to logits $\mathbf{y}_{t+1}$ and to 1-dimensional Q-values: $Q_{\text{acc}}, Q_{\text{div}}$, and $Q_{\text{nov}}$. Diversity and novelty rewards are calculated using the top prediction obtained by the logits. The vector valued Q-value function is set to: $\mathbf{Q} = \bigl[Q_{\text{acc}}, Q_{\text{div}}, Q_{\text{nov}}\bigr]$. SDQL loss is obtained by the scalarization function and SDQL algorithm, and used for training the base model along with the cross-entropy loss.}
\label{fig:architecture}
\end{figure*} 

\section{Model and Training} \label{section:method}
  We cast the task of next item recommendation as a (self-supervised) multi-class classification problem and build a sequential model that receives user-item interaction sequence $\mathbf{x}_{1:t}$ = [$x_1, x_2, ..., x_{t-1}, x_t$] as an input and generates $n$ classification logits $\mathbf{y}_{t+1}=[y_1,y_2,...,y_n] \in \mathbb{R}^n$ ,where $n$ is the number of candidate items. We can then choose the top-$k$ items from $y_{t+1}$ as our recommendation list for timestamp $t+1$. Each candidate item corresponds to a class. 
  
  Typically one can use a generative sequence model $G(\cdot)$ to map the input sequence into a hidden state $\mathbf{s}_t = G(\mathbf{x}_{1:t})$. This serves as a general encoder function. Based on the obtained hidden state, we can utilize a simple decoder to map the hidden state to the classification logits as $y_{t+1} = d(\mathbf{s}_t)$. One can define the decoder function $d$ as a simple fully connected layer or the inner product with candidate item embeddings \cite{hidasi2015session, kang2018self, yuan2020}. In this work, we make use of the fully connected layer. Finally, we train our recommendation model by optimizing the cross-entropy loss $L_s$ based on the logits $y_{t+1}$. Optimization of the cross-entropy loss will push the positive logits to high values, while the items that user did not interact with will be ``penalised'', which will result in a strong negative learning signal. This negative signal is essential for learning in the base model, since the SMORL head provides strong gradients only for positive actions, i.e., top-$1$ items. Furthermore, due to the fact that the sequential recommendation model $G$ has already encoded the input sequence into a latent representation $\mathbf{s}_t$ , we directly use $\mathbf{s}_t$ as the current state for the RL head without the need to introduce a separate RL model. We stack additional fully connected layers to calculate one-dimensional Q-values on top of $G$:
\begin{equation*}
    Q_{z}(\mathbf{s}_t, a_t) = \delta(\mathbf{s}_t\mathbf{H}_z + b_z) = \delta(G(\mathbf{x}_{1:t})\mathbf{H}_z + b_z)
\end{equation*}
where $z \in \{\text{acc}, \text{div}, \text{nov} \}$, $\delta$ denotes the activation function, while $\mathbf{H}_z$ and $b_z$ are learnable parameters of the Q-learning output layer. The SMORL part then stacks computed accuracy, diversity, and novelty Q-values into a vector-valued Q-value function:
\begin{equation}
    \mathbf{Q}(\mathbf{s}_t, a_t) = \bigl[Q_{\text{acc}}(\mathbf{s}_t, a_t), Q_{\text{div}}(\mathbf{s}_t, a_t), Q_{\text{nov}}(\mathbf{s}_t, a_t)\bigr]
\end{equation}

In order to learn vector-valued Q-functions and tackle MORL tasks, \textit{Scalarized Deep Q-learning} (SDQL) \cite{mossalam2016multi} extends the popular DQN algorithm \cite{mnih2013playing}, by introducing a scalarization function $f$. At every time step $t$, Q-network is optimized on the loss $L_{\text{SDQL}}$ computed on a mini-batch of experience tuples $(\mathbf{s}_t, a_t, r_t, s_{t+1})$ obtained from experience buffer $D$:
\begin{equation}
\begin{split}
    L_{\text{SDQL}} &= (f(\mathbf{y}_t^{\text{SDQL}}(\mathbf{s}_t, a_t) - \gamma \mathbf{Q}(\mathbf{s}_{t+1}, a_t)))^2 \\
    &=(\mathbf{w}^\top(\mathbf{y}^{\text{SDQL}}_t(\mathbf{s}_t, a_t) - \gamma \mathbf{Q}(\mathbf{s}_{t+1}, a_t)))^2
    \label{sdql_loss}
\end{split}    
\end{equation}
where $\mathbf{y}_t^{\text{SDQL}}(\mathbf{s}_t, a_t) = \mathbf{r}_t + \gamma \mathbf{Q}'(\mathbf{s}_{t+1}, \argmax_{a'}[\mathbf{w}^\top \mathbf{Q}'(\mathbf{s}_{t+1}, a')])$, and $\mathbf{Q}'$ being the target network. Training towards a fixed target network prevents approximation errors from propagating too quickly from state to state, and sampling experiences to train on (experience replay) increases sample efficiency and reduces correlation between training samples.

When generating recommendations, we still return the top-$k$ items from the supervised head. The SMORL head acts as a regularizer of the base recommendation model $G$ that fine-tunes it by assessing the quality of recommended top item, according to the predefined reward setting and scalarization function $f$, i.e., importance of objectives. 

\subsection{Reinforcing Accuracy}

For the base model $G$ to learn to provide more relevant recommendations, we expand on \cite{xin2020self} and define accuracy reward as
\begin{equation}
r_{\text{acc}}(\mathbf{s}_t, a_t) =  r_{\text{acc}}(a_t) = 1, \quad a_t \,\, \text{is a clicked item}
\end{equation}
From the definition of the reward, the model is rewarded when it matches the next clicked item in the sequence. Xin et. al.~\cite{xin2020self} suggested using the reward for both clicks and purchases. However, in this work, we introduce a method that can be easily extrapolated from e-commerce to other relevant areas of RS. We note that, by reinforcing the relevance of recommended items, one can significantly hinder the user's ability to explore the platform due to the similarity of the recommendations to the user's recent interests. We explore this claim in Section~\ref{experiments}. Therefore, it is crucial for a model to also learn how to recommend diverse sets of items, as well as items that are more probable to never be discovered by the user.

\subsection{Reinforcing Diversity}
\label{divrlhead}

 For the SMORL head to promote diverse sets of recommendations, we first train a GRU4Rec model \cite{hidasi2015session}, and save the embedding layer $\mathbf{E}_\text{div}$. We then freeze the weights of $\mathbf{E}_\text{div}$ to stop further updates of the parameters. We define the reward $r_{\text{div}}$ as 
\begin{equation}\label{reward_div}
r_{\text{div}} = r_{\text{div}}(\mathbf{s}_t, p_t)=1-\text{cos}(l_t, p_t) = 1 - \frac{\mathbf{e}_{l_t}^\top \mathbf{e}_{p_t}}{\|\mathbf{e}_{l_t}\| \|\mathbf{e}_{p_t}\|}
\end{equation}
where $l_t$ is the last clicked item in the session, $p_t$ is a top prediction obtained from self-supervised layer, and $\mathbf{e}_x$ is the embedding of the item $x$, obtained from $\mathbf{E}_\text{div}$. We do not use the embedding of a model that is currently trained for calculation $r_{\text{div}}$. It would be unstable at the beginning of the training process, which would produce unreliable diversity rewards. This reward reinforces diversity across a session of recommendations rather than just over a single slate. Basing the diversity reward system on top prediction $p_t$ and top-$k$ recommendations instead of only the last clicked item $l_t$ was considered, but we observed no improvement in performance. 

\begin{algorithm}[!t]
\SetAlgoLined
\SetKwInOut{Input}{Input}
\SetKwInOut{Output}{Output}
\Input{user-item interaction sequence set $\mathcal{X}$, recommendation model $G$, \newline SMORL head $\mathbf{Q}$ , supervised head S, \newline predefined parameters $\alpha$ and $\mathbf{w}$}
\Output{all parameters in the learning space $\Theta$}
\lstset{numbers=left, numberstyle=\tiny, stepnumber=1, numbersep=5pt}
 Initialize all trainable parameters\\
 Create $G'$, $\mathbf{Q}'$, as copies of $G$ and $\mathbf{Q}$, respectively\\
 \Repeat{$\text{converge}$}{
  Draw a mini-batch of $(\mathbf{x}_{1:t} , a_t)$ from $\mathcal{X}$\\
  $\mathbf{s}_t =G(\mathbf{x}_{1:t}),\mathbf{s}'_t =G'(\mathbf{x}_{1:t})$\\
  $\mathbf{s}_{t+1} = G(\mathbf{x}_{2:t+1}), \mathbf{s}'_{t+1} = G'(\mathbf{x}_{2:t+1})$\\
  Generate random variable $z \in (0, 1)$ uniformly\\
  \eIf{$z < 0.5$}{
   $a^{*} = \argmax_a [\mathbf{Q}(\mathbf{s}_{t+1},a)\cdot \mathbf{w}]$\\
   $\text{pred} = \argmax\,\text{S}(\mathbf{s}_t)$\\
   Set reward $\mathbf{r}_t = \text{stack}(r_{\text{acc}}, r_{\text{div}}, r_{\text{nov}})$ \\
   $L_{\text{SDQL}} = \mathbf{w}^\top(\mathbf{r}_t + \gamma \mathbf{Q}'(\mathbf{s}'_{t+1}, a^*) - \mathbf{Q}(\mathbf{s}_{t}, a_t))^2$\\
   Calculate $L_s$\\
   $L_{\text{SMORL}} = L_s + \alpha L_{\text{SDQL}}$ \\
   Perform updates by $\nabla_{\Theta} L_{\text{SMORL}}$
   }{
   $a^{*} = \argmax_a [\mathbf{Q}'(\mathbf{s}'_{t+1},a)\cdot \mathbf{w}]$\\
   $\text{pred} = \argmax\,\text{S}(\mathbf{s}_t)$\\
   Set reward $\mathbf{r}_t = \text{stack}(r_{\text{acc}}, r_{\text{div}}, r_{\text{nov}})$ \\
   $L_{\text{SDQL}} = (\mathbf{w}^\top(\mathbf{r}_t + \gamma \mathbf{Q}(\mathbf{s}_{t+1}, a^*) - \mathbf{Q}'(\mathbf{s}'_{t}, a_t))^2$\\
   Calculate $L_s$\\
   $L_{\text{SMORL}} = L_s + \alpha L_{\text{SDQL}}$ \\
   Perform updates by $\nabla_{\Theta} L_{\text{SMORL}}$
  }
 }
 Return all parameters in $\Theta$
 \caption{Training Procedure of SMORL}
 \label{smorl_algo}
\end{algorithm}

\subsection{Reinforcing Novelty} \label{novrlhead}
Given an item, a user may have previously seen it in another set of recommendations but chose not to click it, or already encountered it on some other platform. Therefore, in a real-world use case, it is impossible to track the items that a user may have already seen, and to suggest items that are certain to be novel. To address this issue and introduce novelty and serendipity into the set of recommendations, we take a probabilistic approach. Less popular items are more likely to be novel and lead to a more balanced distribution of item popularity. We use binarized item frequency as a novelty reward for our MORL head, which we define as follows:
\begin{equation*}
    r_{\text{nov}} = r_{\text{nov}}(\mathbf{s}_t, p_t) = \begin{cases}
    0.0 &\text{$p_t$ in top $x$\% of most popular items}\\
    1.0 &\text{otherwise}
\end{cases}
\end{equation*}
where $p_t$ is the top predicted item obtained from the self-supervised layer. The choice of $x$ is based on the empirical distribution of the item popularity inferred from the training set, i.e., we set it to the approximate percentile where the long tail starts. Both datasets used in this work have a similar distribution, so we set $x\coloneqq 10$. As can be seen, accuracy reward depends on the next item in the session, while diversity and novelty rewards depends on the top prediction from self-supervised layer. 

\subsection{Scalarized Multi-Objective RL for RS}

Recommendation is by nature a multi-objective problem and, as such, stock self-supervised learning, or even single-objective RL methods, cannot satisfy all desirable (or necessary) goals. 
We integrate the three proposed objectives into a single SMORL method that at each timestamp finds an optimal action that takes into consideration all objectives according to a predefined user utility function, or in this case, according to the configuration of $\mathbf{w}$ from Eq.(\ref{sdql_loss}). SMORL is highly customizable and adaptable to a specific provider's goals - one can define different reward systems that can result in a RS that provides more relevant, novel, diverse, unexpected, or serendipitous recommendations.
The final loss that we optimize is:
\begin{equation}
    L_{SMORL} = L_s + \alpha L_{\text{SDQL}}
    \label{smorl_loss}
\end{equation}
where $L_s$ is a cross-entropy loss, and $\alpha$ is a hyperparameter that enables us to control the influence of SMORL part. In order to enhance the learning stability, we alternately train two copies of learnable parameters. Algorithm \ref{smorl_algo} describes the training procedure of SMORL. It should be noted that after the training is finished, only the self-supervised part of the base model is used to produce recommendations, while the effects with respect to different metrics are observed from the regularization by the SMORL part. 

This training framework can be integrated in existing recommendation models, provided they follow the general architecture discussed earlier. This is the case for most session-based or sequential recommendation models introduced over the last years. In this work, we use the cross-entropy loss for the self-supervised part but other models can incorporate different loss functions~\cite{hidasi2018recurrent, rendle2012bpr}.

In addition, SMORL is a highly modular framework, where one can re-weight and ``deactivate'' specific RL objectives, or add more of them with the help of a carefully designed reward schema. Ultimately, this mechanism allows the RS to focus on providers' specific short-term and long-term goals. However, our experimental results show that models regularized by all three RL objectives perform the best in most cases, with respect to all quality metrics.

\section{Experiments} \label{experiments}
We report the results of our experiments\footnote{The implementation can be found at \url{https://drive.google.com/file/d/1lVeKlajOkZ4n9Rl2VmJvYR9i1aXWkR2j/view?usp=sharing}} 
on two real-world sequential e-commerce datasets. For all base models, we used the self-supervised head to generate recommendations. We address the following research questions:

\textbf{RQ1:} When integrated, does the proposed method increase the performance of the base models?

\textbf{RQ2:} Can we control the balance between accuracy, diversity and novelty?

\textbf{RQ3:} Can we increase the influence of SMORL part by adjusting the intensity of its gradient? 

\subsection{Experimental Settings}
\subsubsection{Datasets:}

RC15\footnote{\hyperlink{https://recsys.acm.org/recsys15/challenge/}{{https://recsys.acm.org/recsys15/challenge/}}} and RetailRocket\footnote{\hyperlink{https://www.kaggle.com/retailrocket/ecommerce-dataset}{https://www.kaggle.com/retailrocket/ecommerce-dataset}}, Table~\ref{datasets}.

\textbf{RC15.} This dataset is based on the RecSys Challange 2015. The dataset is session-based and each session contains a sequence of clicks and purchases\footnote{In this work, we only consider clicks.}. We discard sessions whose length is smaller than 3 and then sample a subset of $200$K sessions.

\textbf{RetailRocket.} This dataset is collected from a real-world e-commerce website. It contains session events of viewing and adding to cart. To keep in line with the RC15 dataset, we treat views as clicks. We remove the items which are interacted less than three times (3), and the sequences whose length is smaller than three (3).

\subsubsection{Quality of Recommendation Metrics}\hfill

\textbf{Accuracy metrics.} Relevance of the recommended item set is usually measured with two metrics: hit ration (HR) and normalized discounted cumulative gain (NDCG). HR@$k$ is a recall-based metric, measuring whether the ground-truth item is in the top-$k$ positions of the recommendation list. We define HR for clicks as:
\begin{equation*}
    \text{HR(click)} = \frac{\text{\#hits among clicks}}{\text{\#clicks}}
\end{equation*}
On the other hand, NDCG is a rank sensitive metric that assigns higher scores to top positions in the recommendation list \cite{jarvelin2002cumulated}

\textbf{Diversity \& Novelty metrics.} Diversity in RS can be viewed at either individual or aggregate level. For example, if the RS was to provide the same set of ten dissimilar items to all users, the recommendation list for each user would be diverse, i.e., it would have high individual diversity. However, the system can only recommend ten items out of the entire item pool and, thus, the aggregate diversity would be negligible. Therefore, in our experiments, we measure aggregate diversity using Coverage@$k$ (CV@$k$), $k\in \{1,5,10,20\}$. More specifically, we measure CV@$k$ on two sets: set of all items and a set of less popular items. Coverage can be computed as percentage of all items (less popular items) covered by all top-$k$ recommendations of the validation or test sequences. 

\textbf{Repetitiveness of Recommendations.} We introduce Repetitiveness (R), a novel metric for evaluating the usefulness of recommendations. We consider this metric a good proxy as to how easily a RS can create a filter bubble, as it measures the per session average of repetitions in the top-$k$ positions of recommendations lists. We measure R@$k$, $k\in \{5,10,20\}$ and define it as:
\begin{equation}
    {R@K = \frac{1}{N} \sum_{i=1}^N \# \text{repetitions in top-$k$ items of session $i$}}  
\end{equation}
where $N$ is the total number of sessions in test (or validation) set. 

\subsubsection{Evaluation Protocols} 

We use 5-fold cross-validation for our performance evaluation, with a ratio of 8:1:1 for training, validation, and testing. We report average performance across all folds.

\begin{table}[!t]
\vspace{-0.75em}
  \centering
  \begin{tabular}{lcc} 
  \toprule
    Dataset & RC15 & RetailRocket
  \\\hline
  \#sequences & 200,000 & 195,523\\
  \#items & 26,702 & 70,852 \\
  \#clicks & 1,110,965 & 1,176,680 \\
  \#purchase & 43,946 & 57,269 \\
  \bottomrule
  \end{tabular}
  \caption{Dataset statistics.}
  \label{datasets}
\end{table}

\subsubsection{Baselines}

We integrated SMORL in four state-of-the-art (generative) sequential recommendation models:
\begin{itemize}
    \item GRU4Rec \cite{hidasi2015session}: This method uses a GRU to model the input sequences. The final hidden state of the GRU4Rec is treated as the latent representation for the input sequence.
    \item Caser \cite{tang2018personalized}: This recently introduced CNN-based method captures sequential signals by applying convolution operations on the embedding matrix of previous items.
    \item NextItNet \cite{yuan2020}: This method enhances Caser by using dilated CNN to enlarge the receptive field and residual connection to increase the network depth.
    \item SASRec \cite{kang2018self}: This baseline is motivated from self-attention and uses the Transformer \cite{vaswani2017attention} architecture to encode sequences of user-item interactions. The output of the Transformer encoder is treated as the latent representation.
\end{itemize}

\subsubsection{Parameter settings} 

For both datasets the input sequences comprise of the last $10$ items before the target timestamp. If the sequence length is less than $10$, we complement it with a padding item. We train all models with the Adam optimizer \cite{kingma2014adam}. The mini-batch size is set as 256. The learning rate is set as 0.01 for RC15 and 0.005 for RetailRocket. We evaluate on the validation set every $5,000$ batches of updates on RC15, and every $10,000$ batches of updates on RetailRocket. To ensure a fair comparison, the item embedding size is set as $64$ for all models. For the GRU4Rec model, the size of the hidden state is set as $64$. For Caser, we use one vertical convolution filter and $16$ horizontal filters, whose heights are set from $\{2,3,4\}$. The drop-out ratio is set as $0.1$. For NextItNet, we use the same parameters reported by authors. For SASRec, the number of heads in self-attention is set as $1$, according to its original paper \cite{kang2018self}. We set the discount factor $\gamma$ to $0.5$, as recommended by ~\citet{xin2020self}. 

  \setlength{\tabcolsep}{1.5pt}
  \begin{table*}
  \vspace{-0.75em}
  \centering
    \begin{tabular}{lccccccccccccccc}
      \toprule
      \multirow{2}{*}{Models} & \multicolumn{4}{c}{accuracy} & \multicolumn{4}{c}{diversity} & \multicolumn{4}{c}{novelty} & \multicolumn{3}{c}{repetitiveness}
      \\\cmidrule(rl){2-5} \cmidrule(rl){6-9} \cmidrule(rl){10-13} \cmidrule(rl){14-16}
      & HR@10 & NG@10 & HR@20 & NG@20 & CV@1 & CV@5 & CV@10 & CV@20 & CV@1 & CV@5 & CV@10 & CV@20 & R@5 & R@10 & R@20
      \\\hline
      GRU & 0.3793 & 0.2279 & 0.4581 & 0.2478 & 0.2481 & 0.4330 & 0.5188 & 0.5942 & 0.1777 & 0.3707 & 0.4654 & 0.5492 & 12.11 & 25.63 & 53.24 \\
      GRU-SQN & 0.3946 & 0.2394 & 0.4741 & 0.2587 & 0.2406 & 0.4025 & 0.4710 & 0.5364 & 0.1656 & 0.3363 & 0.4122 & 0.4849  & 12.20 & 25.81 & 53.47 \\
      GRU-SMORL & \textbf{0.4007} & \textbf{0.2433} & \textbf{0.4793} & \textbf{0.2632} & \textbf{0.2825} & \textbf{0.4758} & \textbf{0.5577} & \textbf{0.6334} & \textbf{0.2086} & \textbf{0.4176} & \textbf{0.5086} & \textbf{0.5927} & \textbf{11.29} & \textbf{23.81} & \textbf{48.88}
      \\\hline
      Caser & 0.3593 & 0.2177 & 0.4371 & 0.2372 & 0.2631 & 0.4349 & 0.5019 & 0.5608	 & 0.1912 & 0.3724 & 0.4466 & 0.5120 & 14.38 & 29.65 & 60.73 \\
      Caser-SQN & \textbf{0.3668} & 0.2223 & \textbf{0.4448} & \textbf{0.2420} & 0.2154 & 0.3525 & 0.4057 & 0.4557 & 0.1411 & 0.2810 & 0.2154 & 0.3953 & 14.45 & 29.79 & 60.82\\
      Caser-SMORL & 0.3664 & \textbf{0.2224} & 0.4425 & 0.2417 & \textbf{0.3174} & \textbf{0.5157} & \textbf{0.5944} & \textbf{0.6685} & \textbf{0.2476} & \textbf{0.4621} & \textbf{0.5495} & \textbf{0.6316} & \textbf{13.77} & \textbf{28.56} & \textbf{58.52}
      \\\hline
      NtItNet & 0.3885 & 0.2332 & 0.4684 & 0.2535 & 0.2950 & 0.4914 & 0.5705 & 0.6427 & 0.2313 & 0.4354 & 0.5228 & 0.6030 & 10.03 & 22.02 & 46.84 \\
      NtItNet-SQN & 0.4083 & 0.2492 & 0.4878 & 0.2693 & 0.2737 & 0.4572 & 0.5183 & 0.5715 & 0.2082 & 0.3975 & 0.4649 & 0.5239 & 10.19 & 22.32 & 47.26\\
      NtItNet-SMORL & \textbf{0.4116} & \textbf{0.2505} & \textbf{0.4898} & \textbf{0.2703} & \textbf{0.3385} & \textbf{0.5639} & \textbf{0.6518} & \textbf{0.7283} & \textbf{0.2720} & \textbf{0.5156} & \textbf{0.6131} & \textbf{0.6981} & \textbf{9.97} & \textbf{21.73} & \textbf{45.49} 
      \\\hline
      SASRec & 0.4257 & 0.2599 & 0.5053 & 0.2801 & 0.2971 & 0.5208 & 0.6046 & 0.6792 & 0.2298 & 0.4679 & 0.5607 & 0.6436 & 10.62 & 23.24 & 49.28\\
      SASRec-SQN & 0.4288 & 0.2630 & 0.5073 & 0.2829 & 0.2701 & 0.4527 & 0.5194 & 0.5755 & 0.2018 & 0.3922 & 0.4660 & 0.5283 & 10.94 & 23.85 & 50.79 \\
      SASRec-SMORL & \textbf{0.4315} & \textbf{0.2651} & \textbf{0.5104} & \textbf{0.2851} & \textbf{0.3380} & \textbf{0.5755} & \textbf{0.6508} & \textbf{0.7158} & \textbf{0.2698} & \textbf{0.5285} & \textbf{0.6120} & \textbf{0.6842} & \textbf{10.38} & \textbf{22.79} & \textbf{48.48}
      \\\bottomrule
    \end{tabular}
    \caption{Recommendation performance on RC15 dataset. NG is NDCG. CV is Coverage. Boldface denotes highest score.}
    \label{rc15_results}
    \end{table*}
    
  \begin{table*}
    \vspace{-0.75em}
  \centering
    \begin{tabular}{lccccccccccccccc}
      \toprule
       \multirow{2}{*}{Models} & \multicolumn{4}{c}{accuracy} & \multicolumn{4}{c}{diversity} & \multicolumn{4}{c}{novelty} & \multicolumn{3}{c}{repetitiveness}
      \\\cmidrule(rl){2-5} \cmidrule(rl){6-9} \cmidrule(rl){10-13} \cmidrule(rl){14-16}
      & HR@10 & NG@10 & HR@20 & NG@20 & CV@1 & CV@5 & CV@10 & CV@20 & CV@1 & CV@5 & CV@10 & CV@20 & R@5 & R@10 & R@20
      \\\hline
      GRU & 0.2673 & 0.1878 & 0.3082 & 0.1981 & 0.2439 & 0.4695 & 0.5699 & 0.6632 & 0.1837 & 0.4139 & 0.5238 & 0.6267 & 14.25 & 29.44 & 60.59\\
      GRU-SQN & 0.2967 & 0.2094 & 0.3406 & 0.2205 & 0.2180 & 0.4114 & 0.4975 & 0.5763 & 0.1526 & 0.3489 & 0.4430 & 0.5299 & 14.62 & 30.19 & 62.22 \\
      GRU-SMORL & \textbf{0.3060} & \textbf{0.2103} & \textbf{0.3535} & \textbf{0.2224} & \textbf{0.2796} & \textbf{0.5369} & \textbf{0.6419} & \textbf{0.7353} & \textbf{0.2154} & \textbf{ 0.4871} & \textbf{0.6029} & \textbf{0.7064} & \textbf{13.53} & \textbf{28.02} & \textbf{57.89}
      \\\hline
      Caser & 0.2302 & 0.1675 & 0.2628 & 0.1758 & 0.2327 & 0.4379 & 0.5133 & 0.5718 & 0.1643 & 0.3773 & 0.4605 & 0.5252 & 16.16 & 33.24 & 68.39\\
      Caser-SQN & 0.2454 & 0.1778 & 0.2803 & 0.1867 & 0.2088 & 0.3880 & 0.4511 & 0.5021 & 0.1387 & 0.3219 & 0.3914 & 0.4479 & 16.88 & 34.50 & 70.58\\
      Caser-SMORL & \textbf{0.2657} & \textbf{0.1898} & \textbf{0.3052} & \textbf{0.1998} & \textbf{0.2855} & \textbf{0.5411} & \textbf{0.6324} & \textbf{0.7138} & \textbf{0.2224} & \textbf{0.4917} & \textbf{0.5925} & \textbf{0.6827 } &  \textbf{15.90} & \textbf{32.47} & \textbf{66.76}
      \\\hline
      NtItNet & 0.3007 & 0.2060 & 0.3506 & 0.2186 & 0.2867 & 0.5113 & 0.6033 & 0.6837 & 0.2305 & 0.4595 & 0.5605 & 0.6495 & 12.25 & 25.76 & 54.00\\
      NtItNet-SQN & 0.3129 & 0.2150 & 0.3586 & 0.2266 & 0.2802 & 0.5255 & 0.6077 & 0.6750 & 0.2184 & 0.4747 & 0.5651 & 0.6395 & 12.27 & 25.93 & 54.47 \\
      NtItNet-SMORL & \textbf{0.3183} & \textbf{0.2222} & \textbf{0.3659} & \textbf{0.2342} & \textbf{0.3429} & \textbf{0.6335} & \textbf{0.7351} & \textbf{0.8129} & \textbf{0.2800} & \textbf{0.5938} & \textbf{0.7062} & \textbf{0.7924} & \textbf{10.92} & \textbf{22.89} & \textbf{47.73}
      \\\hline
      SASRec & 0.3085 & 0.2107 & 0.3572 & 0.2227 & 0.2767 & 0.5305 & 0.6300 & 0.7149 & 0.2171 & 0.4806 & 0.5899 & 0.6838 & 15.67 & 32.27 & 66.07\\
      SASRec-SQN & 0.3302 & 0.2279 & 0.3803 & 0.2406 & 0.2393 & 0.4617 & 0.5490 & 0.6254 & 0.1753 & 0.4040 & 0.5001 & 0.5847 & 15.60 & 32.20 & 66.10 \\
      SASRec-SMORL & \textbf{0.3521} & \textbf{0.2477} & \textbf{0.4028} & \textbf{0.2605} & \textbf{0.3037} & \textbf{0.5724} & \textbf{0.6672} & \textbf{0.7476} & \textbf{0.2366} & \textbf{0.5261} & \textbf{0.6311} & \textbf{0.7202} & \textbf{12.58} & \textbf{26.69} & \textbf{56.14}
      \\\bottomrule
    \end{tabular}
    \caption{Recommendation performance on RetailRocket dataset. NG is NDCG. CV is Coverage. Boldface denotes highest score.}
    \label{retail_rocket_results}
    \end{table*}
    
\begin{figure*}[!t]
    \captionsetup[subfloat]
    {}
    \centering
    \subfloat[Caser]{
    \label{xxxxxx}
    \includegraphics[clip=true, trim=0 0 11 0, width=0.25\textwidth]{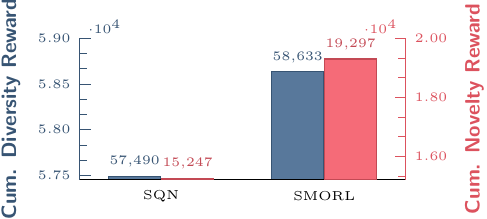}}
    \hspace{0.1mm}
    \subfloat[GRU]{%
    \label{xxxxxx}
    \includegraphics[clip=true, trim=11 0 11 0, width=0.23\textwidth]{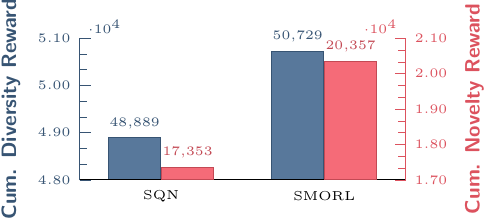}}
    \hspace{0.1mm}
    \subfloat[NextItNet]{%
    \label{xxxxxx}
    \includegraphics[clip=true, trim=11 0 11 0, width=0.23\textwidth]{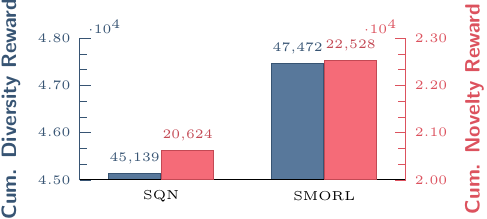}}
    \hspace{0.1mm}
    \subfloat[SASRec]{%
    \label{xxxxxx}
    \includegraphics[clip=true, trim=11 0 0 0, width=0.25\textwidth]{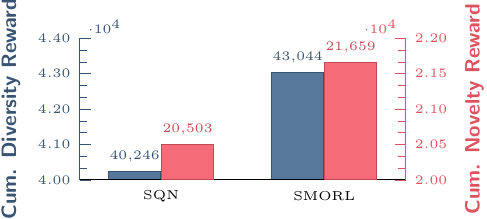}}
    \vspace{-0.75em}
    \caption{Comparison of cumulative rewards on RC15 dataset with base models regularized by SMORL and SQN frameworks.}
    \label{figure:cumulative_rewards}
\end{figure*}

\subsection{Performance Comparison (RQ1)}
For both datasets, the SQN method~\cite{xin2020self} outperforms the baselines with respect to recommending relevant items to the users. 
However, by increasing the accuracy of the baseline model, it causes it to ``drift'' from diversity and novelty. This results in a substantial decrease (up to 20\%) of coverage metrics for the baseline model, both on all and less popular items. Together with this fact, increased repetitiveness of recommendations suggests that reinforcing accuracy alone may hinder significantly the perceived quality of experience. Furthermore, it is evident that one should simultaneously optimize the model towards diversity and novelty to achieve a balance between opposing metrics. In Table \ref{rc15_results} and Table \ref{retail_rocket_results}, we see that by using the SMORL method we not only obtain a balance between accuracy, diversity and novelty, but we consistently outperform the corresponding baselines across all metrics and, to some extent, we also improve their accuracy power. The increase in diversity and novelty is up to 20\% relative to the baseline model, and up to 40\% relative to the SQN model. Increases in the accuracy of the baseline models can be attributed to most users having diverse interests that cannot be satisfied by the recommendations produced by an RS~\cite{anderson2020algorithmic}. 
Figure~\ref{figure:cumulative_rewards} displays the difference in cumulative diversity and novelty rewards obtained on the RC15 test set. When a base model is trained with the SMORL framework, we note a significant increase in the cumulative diversity and novelty rewards. Also, the results in Tables \ref{rc15_results} and  \ref{retail_rocket_results} suggest that reinforcing diversity and novelty introduces a notable improvement in these metrics, which are highly correlated with perceived quality of experience and engagement.

\begin{figure*}[!t]
    \captionsetup[subfloat]
    {}
    \centering
    \subfloat[Accuracy]{
    \label{w_configs_acc}
    \includegraphics[width=0.24\textwidth]{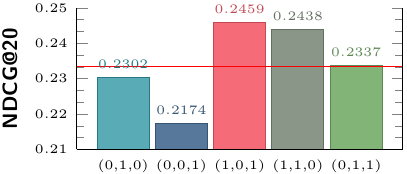}}
    \subfloat[Diversity]{%
    \label{w_configs_div}
    \includegraphics[width=0.24\textwidth]{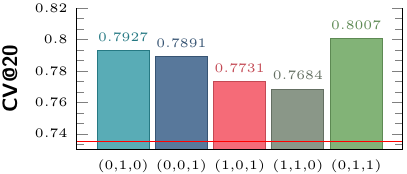}}
    \subfloat[Novelty]{%
    \label{w_configs_nov}
    \includegraphics[width=0.24\textwidth]{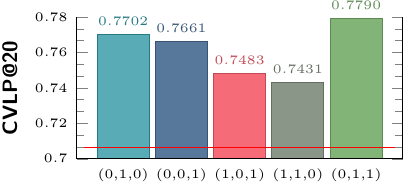}}
    \subfloat[Repetitiveness]{%
    \label{w_configs_repet}
    \includegraphics[width=0.24\textwidth]{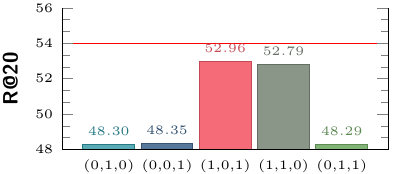}}
    \vspace{-0.75em}
    \caption{Performance comparison when reinforcing a subset of objectives (achieved by using different configurations of $\mathbf{w}$ from Eq.(\ref{equation:weight_conf})). Red lines denote relevant metrics of the stock NextItNet model - not trained using the SMORL4RS framework.}
    \label{w_configs}
\end{figure*}

\subsection{Reinforcing a Subset of Objectives (RQ2)} \label{objectives_weights}

One of the advantages of using SMORL is its objective-balancing capability, which works by re-weighting the objectives using different configurations of $\mathbf{w}$'s in Eq.(\ref{sdql_loss}). In our setting, the first entry of $\mathbf{w}$ corresponds to the strength of accuracy objective, the second to diversity, and the third to novelty objective. We conduct experiments with the following configurations of the parameter $\mathbf{w}$:
\begin{equation} 
\label{equation:weight_conf}
    \mathbf{w} \in \{(0,1,0), (0,0,1), (0,1,1), (1,1,0), (1,0,1)\}
\end{equation}
Here, we aim to demonstrate the difference in performance when reinforcing a subset of three important objectives. We do not include $\mathbf{w} = (1,0,0)$ in this analysis, since SMORL becomes equivalent to SQN method from \cite{xin2020self} and our results show exactly the same behaviour across all models. 

The objectives that we address in this work have a complex relationship. For example, relevance and diversity at the beginning of the training process are correlated, i.e., more diverse recommendations produce more relevant recommendations, while their correlation becomes negative as the training progresses. Diversity and novelty are intertwined objectives, e.g., a diverse set of recommended items is more likely to contain novel items. On the other hand, the popularity of items follows a power distribution and, therefore, less popular items make up to 90\% of the dataset, which means that items likely to be novel are inherently diverse. Given that the proposed method is not a pure MORL model, but rather a regularizer that forces the base model to capture different (and often competing) objectives, the intricacies of optimizing and balancing multiple objectives pose a significant research challenge. In this section, our goal is to demonstrate that we can control how much influence each objective has, and not how to find an ideal balance. With the ability of control, many engineering possibilities arise, such as deploying multiple SMORL4RS agents and deciding in an online fashion if a user should receive recommendations from an agent that is optimized towards novelty, diversity, or accuracy.

\begin{figure}[!t]
    \captionsetup[subfloat]
    {}
    \centering
    \subfloat[]{
    \label{fig:nextitnet_alpha_retail_rc15}
    \includegraphics[clip=true, trim=0 0 10 0, width=0.235\textwidth]{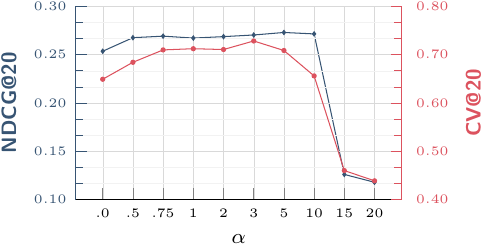}}
    \subfloat[]{%
    \label{fig:nextitnet_alpha_retail_rocket}
    \includegraphics[clip=true, trim=10 0 0 0, width=0.235\textwidth]{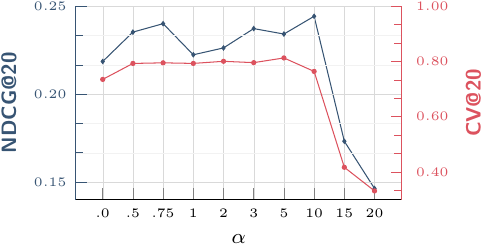}}
    \vspace{-0.75em}
    \caption{NextItNet with different intensity of SMORL gradient on the RC15 (\ref{fig:nextitnet_alpha_retail_rc15}) and RetailRocket (\ref{fig:nextitnet_alpha_retail_rocket}) datasets}
    \label{fig:nextitnet_alpha}
    \vspace{-0.5cm}
\end{figure}

Figure~\ref{w_configs} shows the comparison of NextItNet-SMORL model regularized by the SMORL agent that uses mentioned weight configurations $\mathbf{w}$ on RetailRocket dataset, while similar behaviour can be observed for the RC15 dataset and other models. More specifically, Figure~\ref{w_configs_acc} indicates that if we regularize the model only towards novelty, we will sacrifice its ability to recommend relevant items. This phenomenon is also present if we only reinforce towards diversity, but the drop in NDCG@20 metric is not as notable. On the other hand, if we optimize jointly towards diversity and novelty, we do not observe a drop in the accuracy of the base model. Additionally, if we include the accuracy objective to any of the two other, we observe an increase in the relevant metric. From Figures \ref{w_configs_div} and \ref{w_configs_nov}, we note that including the accuracy objective comes at the cost of diversity and novelty, while combined optimization towards diversity and novelty produce the best results with respect to these metrics. Similarly, by including the accuracy objective, we increase the repetitiveness compared to the NextItNet-SMORL model that optimizes towards a combination of diversity and novelty. 

\subsection{Gradient Intensity Investigation (RQ3)} 
Across all base models and both datasets, the SDQL loss is dominated by self-supervised loss, which suggests that the optimization of parameter $\alpha$ from Eq.(\ref{smorl_loss}) might improve the effect of SMORL part on the base model. Figure \ref{fig:nextitnet_alpha} shows the behaviour of NextItNet-SMORL model with respect to NDCG@20 and CV@20 metrics on both datasets when we change the intensity of SDQL gradient. As expected, when multiplying SDQL with $\alpha < 1$, the effects are decreased and we do not improve dramatically compared to the base model. Increase in both metrics can be seen for $\alpha\in \{1,2,3,5,10\}$, with the best balance obtained for $\alpha=5$. For higher values of $\alpha$, we observe a notable drop in quality due to the loss of gradient signal obtained from the self-supervised loss, which indicates that it is necessary to have a self-supervised part to learn basic ranking. Similar analysis can be made for RC15 dataset. 

The optimal value of the $\alpha$ parameter is equal to 1 for most cases - SASRec on RC15 dataset, GRU4Rec on RetailRocket, and Caser on the RetailRocket dataset. However, for GRU4Rec and Caser on RC15, the optimal value is equal to 0.75, for SASRec and NextItNet on RC15 to 3, while for SASRec on RetailRocket is equal to 10. Hence for real-world use-cases, when datasets usually contain millions of items, higher values of $\alpha$ might be  optimal. More complex models, such as NextItNet and SASRec require higher value of $\alpha$. 

 \section{Conclusions \& Future Work}

 We first formalized the next item recommendation task and presented it as a Multi-Objective MDP task. The SMORL method acts as a regularizer for introducing desirable properties into the recommendation model, specifically to achieve a balance between relevance, diversity and novelty of recommendations. We integrated SMORL with four state-of-the-art recommendation models and conducted experiments on two real-world e-commerce datasets. Our experimental findings demonstrate that the joint optimization of three conflicting objectives is essential for improving metrics that are strongly correlated with user satisfaction, while also preserving content relevance. Future work brings vast possibilities for exploring the use of SMORL paradigm in the setting of RS, and it will include further experiments with different objectives and application of SMORL in different areas, such as music platforms. Also, the joint optimization of supervised and SDQL loss is a research problem on its own. Finally, we plan on exploring the use of non-linear and personalized scalarization functions.

\balance{}
\newpage
\bibliographystyle{ACM-Reference-Format}
\bibliography{main}
\end{document}